\documentclass{llncs}
\usepackage{makeidx}  

\usepackage{graphicx}
\graphicspath{{pics/}{figs/}}

\usepackage{floatrow}
\newfloatcommand{capbtabbox}{table}[][\FBwidth]

\begin{document}
\frontmatter          
\pagestyle{empty}  
\mainmatter              
%
\title{Tversky loss function for image segmentation using 3D fully convolutional deep networks}
\titlerunning{Short Title}  
\author{Seyed Sadegh Mohseni Salehi\inst{1,2}\thanks{Corresponding author: S.S.M.Salehi (email: ssalehi@ece.neu.edu).} \and Deniz Erdogmus\inst{1}  \and Ali Gholipour\inst{2} }
%

\tocauthor{Rasmus Larsen (Technical University of Denmark),
author2 (Affiliation of Author2), author3 (Affiliation of
Author3), }
\institute{Electrical and Computer Engineering Department, Northeastern University
  \and
  Radiology Department, Boston Children's Hospital; and Harvard Medical School, Boston, MA, 02115,
}


\maketitle              

\begin{abstract}
Fully convolutional deep neural networks carry out excellent potential for fast and accurate image segmentation. One of the main challenges in training these networks is data imbalance, which is particularly problematic in medical imaging applications such as lesion segmentation where the number of lesion voxels is often much lower than the number of non-lesion voxels. Training with unbalanced data can lead to predictions that are severely biased towards high precision but low recall (sensitivity), which is undesired especially in medical applications where false negatives are much less tolerable than false positives. Several methods have been proposed to deal with this problem including balanced sampling, two step training, sample re-weighting, and similarity loss functions. In this paper, we propose a generalized loss function based on the Tversky index to address the issue of data imbalance and achieve much better trade-off between precision and recall in training 3D fully convolutional deep neural networks. Experimental results in multiple sclerosis lesion segmentation on magnetic resonance images show improved $F_2$ score, Dice coefficient, and the area under the precision-recall curve in test data. Based on these results we suggest Tversky loss function as a generalized framework to effectively train deep neural networks. 
\end{abstract}
\vspace{-0.5cm}
\section{Introduction}
\label{sec:intro}
Deep convolutional neural networks have attracted enormous attention in medical image segmentation as they have shown superior performance compared to conventional methods in several applications. This includes automatic segmentation of brain lesions \cite{brosch2015deep,kamnitsas2017efficient}, tumors \cite{havaei2017brain,pereira2016brain,wachinger2017deepnat}, and neuroanatomy \cite{moeskops2016automatic,zhang2015deep,chen2017voxresnet}, using voxelwise network architectures \cite{moeskops2016automatic,havaei2017brain,salehi2017auto}, and more recently using 3D voxelwise networks \cite{chen2017voxresnet,kamnitsas2017efficient}, and fully convolutional networks (FCNs) \cite{cciccek20163d,milletari2016v,salehi2017auto}. Compared to voxelwise methods, FCNs are fast in testing and training, and use the entire samples to learn local and global image features. On the other hand, voxelwise networks may use a subset of samples to reduce data imbalance issues and increase efficiency.

Data imbalance is a common issue in medical image segmentation. For example in lesion detection the number of non-lesion voxels is typically $>500$ times larger than the number of diagnosed lesion voxels. Without balancing the labels the learning process may converge to local minima of a sub-optimal loss function, thus predictions may strongly bias towards non-lesion tissue. The outcome will be high-precision, low-recall segmentations. This is undesired especially in computer-aided diagnosis or clinical decision support systems where high sensitivity (recall) is a key characteristic of an automatic detection system.

A common approach to account for data imbalance, especially in voxelwise methods, is to extract equal training samples from each class \cite{valverde2017improving}. The downsides of this approach are that it does not use all the information content of the images and may bias towards rare classes. Hierarchical training \cite{cirecsan2013mitosis,wachinger2017deepnat,valverde2017improving} and retraining \cite{havaei2017brain} have been proposed as alternative strategies but they can be prone to overfitting and sensitive to the state of the initial classifiers \cite{kamnitsas2017efficient}. Recent training methods for FCNs resorted to loss functions based on sample re-weighting \cite{brosch2015deep,kamnitsas2017efficient,long2015fully,ronneberger2015u,shelhamer2017fully}, where lesion regions, for example, are given more importance than non-lesion regions during training. In the re-weighting approach, to balance the training samples between classes, the total cost is calculated by computing the weighted mean of each class. The weights are inversely proportional to the probability of each class appearance, i.e. higher appearance probabilities lead to lower weights. Although this approach works well for some relatively unbalanced data like brain extraction \cite{salehi2017auto} and tumor detection \cite{pereira2016brain}, it becomes difficult to calibrate and does not perform well for highly unbalanced data such as lesion detection. To eliminate sample re-weighting, Milletari et. al. proposed a loss function based on the Dice similarity coefficient \cite{milletari2016v}.

The Dice loss layer is a harmonic mean of precision and recall thus weighs false positives (FPs) and false negatives (FNs) equally. To achieve a better trade-off between precision and recall (FPs vs. FNs), we propose a loss layer based on the Tversky similarity index \cite{tversky1977features}. Tversky index is a generalization of the Dice similarity coefficient and the $F_\beta$ scores. We show how adjusting the hyperparameters of this index allow placing emphasis on false negatives in training a network that generalizes and performs well in highly imbalanced data as it leads to high sensitivity, Dice, $F_2$ score, and the area under the precision-recall (PR) curve \cite{boyd2013area} in the test set. To this end, we adopt a 3D FCN, based on the U-net architecture, with a Tversky loss layer, and test it in the challenging multiple sclerosis lesion detection problem on multi-channel MRI \cite{commowick2016msseg,valverde2017improving}. The ability to train a network for higher sensitivity (recall) in the expense of acceptable decrease in precision is crucial in many medical image segmentation tasks such as lesion detection.
\vspace{-0.3cm}
\section{Method}
\vspace{-0.3cm}
\subsection{Network architecture}
\begin{figure}
    \centering
    \includegraphics[width=\textwidth]{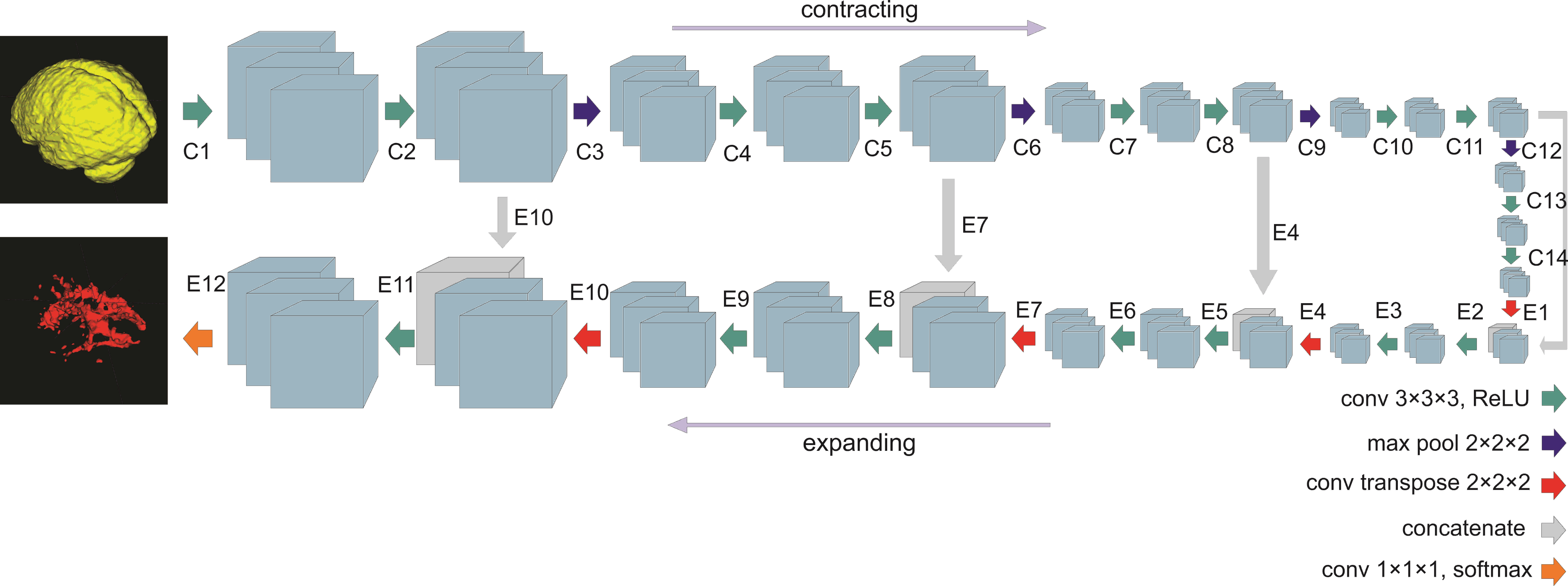}
    \caption{The 3D U-net style architecture; The complete description of the input and output size of each level is presented in Table~S1 in the supplementary material.}
    \label{fig:Net}
\end{figure}

We design and evaluate our 3D fully convolutional network~\cite{long2015fully,shelhamer2017fully} based on the U-net architecture~\cite{ronneberger2015u}. To this end, we develop a 3D U-net based on Auto-Net~\cite{salehi2017auto} and introduce a new loss layer based on the Tversky index. This U-net style architecture, which has been designed to work with very small number of training images, is shown in Figure~\ref{fig:Net}. It consists of a contracting path (to the right) and an expanding path (to the left). To learn and use local information, high-resolution 3D features in the contracting path are concatenated with upsampled versions of global low-resolution 3D features in the expanding path. Through this concatenation the network learns to use both high-resolution local features and low-resolution global features.

The contracting path contains padded $3\times3\times3$ convolutions followed by ReLU non-linear layers. A $2\times2\times2$ max pooling operation with stride 2 is applied after every two convolutional layers. After each downsampling by the max pooling layers, the number of features is doubled. In the expanding path, a $2\times2\times2$ transposed convolution operation is applied after every two convolutional layers, and the resulting feature map is concatenated to the corresponding feature map from the contracting path. At the final layer a $1\times1\times1$ convolution with softmax output is used to reach the feature map with a depth equal to the number of classes (lesion or non-lesion tissue), where the loss function is calculated. The size of the network layers is shown in Table~S1 in the supplementary material.

\vspace{-0.3cm}
\subsection{Tversky loss layer}
The output layer in the network consists of $c$ planes, one per class ($c=2$ in lesion detection). We applied softmax along each voxel to form the loss. Let $P$ and $G$ be the set of predicted and ground truth binary labels, respectively. The Dice similarity coefficient $D$ between two binary volumes is defined as: 
\begin{equation}
    D(P,G) = \frac{2|PG|}{|P|+|G|}
\end{equation}
If this is used in a loss layer in training \cite{milletari2016v}, it weighs FPs and FNs (precision and recall) equally. In order to weigh FNs more than FPs in training our network for highly imbalanced data, where detecting small lesions is crucial, we propose a loss layer based on the Tversky index \cite{tversky1977features}. The Tiversky index is defined as:
\begin{equation}
    S(P,G;\alpha,\beta) = \frac{|PG|}{|PG|+\alpha|P\setminus G|+\beta|G\setminus P|}
\end{equation}
\noindent where $\alpha$ and $\beta$ control the magnitude of penalties for FPs and FNs, respectively. 

To define the Tversky loss function we use the following formulation:
\begin{equation}
    T(\alpha,\beta) = \frac{\sum_{i=1}^Np_{0i}g_{0i}}{\sum_{i=1}^Np_{0i}g_{0i} + \alpha \sum_{i=1}^Np_{0i}g_{1i} + \beta \sum_{i=1}^Np_{1i}g_{0i}}
    \label{tversky}
\end{equation}
\noindent where in the output of the softmax layer, the $p_{0i}$ is the probability of voxel $i$ be a lesion and $p_{1i}$ is the probability of voxel $i$ be a non-lesion. Also, $g_{0i}$ is 1 for a lesion voxel and 0 for a non-lesion voxel and vice verse for the $g_{1i}$. The gradient of the loss in Equation \ref{tversky} with respect to $p_{0i}$ and $p_{1i}$ can be calculated as:
\small
\begin{equation}
        \frac{\partial T}{\partial p_{0i}} = 2\frac{g_{0j}(\sum_{i=1}^Np_{0i}g_{0i} + \alpha \sum_{i=1}^Np_{0i}g_{1i} + \beta \sum_{i=1}^Np_{1i}g_{0i}) - (g_{0j} + \alpha g_{1j})\sum_{i=1}^Np_{0i}g_{0i}}{(\sum_{i=1}^Np_{0i}g_{0i} + \alpha \sum_{i=1}^Np_{0i}g_{1i} + \beta \sum_{i=1}^Np_{1i}g_{0i})^2}
\end{equation}

\begin{equation}
        \frac{\partial T}{\partial p_{1i}} = -\frac{\beta g_{1j} \sum_{i=1}^Np_{0i}g_{0i}}{(\sum_{i=1}^Np_{0i}g_{0i} + \alpha \sum_{i=1}^Np_{0i}g_{1i} + \beta \sum_{i=1}^Np_{1i}g_{0i})^2}
\end{equation}
\normalsize
Using this formulation we do not need to balance the weights for training. Also by adjusting the hyperparameters $\alpha$ and $\beta$ we can control the trade-off between false positives and false negatives. It is noteworthy that in the case of $\alpha = \beta = 0.5$ the Tversky index simplifies to be the same as the Dice coefficient, which is also equal to the $F_1$ score. With $\alpha = \beta = 1$, Equation 2 produces Tanimoto coefficient, and setting $\alpha+\beta=1$ produces the set of $F_\beta$ scores. Larger $\beta$s weigh recall higher than precision (by placing more emphasis on false negatives). We hypothesize that using higher $\beta$s in our generalized loss function in training will lead to higher generalization and improved performance for imbalanced data; and effectively helps us shift the emphasis to lower FNs and boost recall.
\vspace{-0.3cm}
\subsection{Experimental design}
We tested our FCN with Tversky loss layer to segment multiple sclerosis (MS) lesions \cite{commowick2016msseg,valverde2017improving}. T1-weighted, T2-weighted, and FLAIR MRI of 15 subjects were used as input, where we used two-fold cross-validation for training and testing. Images of different sizes were all rigidly registered to a reference image at size $128 \times 224 \times 256$. Our 3D-Unet was trained end-to-end. Cost minimization on 1000 epochs was performed using ADAM optimizer~\cite{kingma2014adam} with an initial learning rate of 0.0001 multiplied by 0.9 every 1000 stjpg. The training time for this network was approximately 4 hours on a workstation with Nvidia Geforce GTX1080 GPU.

The test fold MRI volumes were segmented using feedforward through the network. The output of the last convolutional layer with softmax non-linearity consisted of a probability map for lesion and non-lesion tissues. Voxels with computed probabilities of 0.5 or more were considered to belong to the lesion tissue and those with probabilities $<0.5$ were considered non-lesion tissue.
\vspace{-0.3cm}
\subsection{Evaluation metrics}
To evaluate the performance of the networks and compare them against state-of-the-art in MS lesion segmentation, we report Dice similarity coefficient (DSC):
\begin{equation}
    DSC = \frac{2\left | P\cap R \right |}{\left | P \right |+\left | R \right |} = \frac{2TP}{2TP+FP+FN}
\end{equation}
where $P$ and $R$ are the predicted and ground truth labels, respectively; and $TP$, $FP$, and $FN$ are the true positive, false positive, and false negative rates, respectively. We also calculate and report specificity, $\frac{TN}{TN+FP}$, and sensitivity, $\frac{TP}{TP+FN}$, and the $F_2$ score as a measure that is commonly used in applications where recall is more important than precision (as compared to $F_1$ or DSC):
\begin{equation}
    F_2 = \frac{5TP}{5TP+4FN+FP}
\end{equation}
To critically evaluate the performance of the detection for the highly unbalanced (skewed) dataset, we use the Precision-Recall (PR) curve (as opposed to the receiver-operator characteristic, or ROC, curve) as well as the area under the PR curve (the APR score)~\cite{boyd2013area,davis2006relationship,fawcett2006introduction}. For such skewed datasets, the PR curves and APR scores (on test data) are preferred figures of algorithm performance.
\vspace{-0.3cm}
\section{Results}
\vspace{-0.3cm}
To evaluate the effect of Tversky loss function and compare it with Dice in lesion segmentation, we trained our FCN with different $\alpha$ and $\beta$ values. The performance metrics (on the test set) have been reported in Table~\ref{table:res}. The results show that 1) the balance between sensitivity and specificity was controlled by the parameters of the loss function; and 2) according to all combined test measures, the best results were obtained from the FCN trained with $\beta=0.7$, which performed much better than the FCN trained with the Dice loss layer corresponding to $\alpha=\beta=0.5$.

 

\vspace{-0.3cm}
\begin{table*}[h!]
\small
\centering
 \begin{tabular}{|c||c|c|c|c|c|c|c|} 
 \hline
Penalties & DSC & Sensitivity & Specificity & $F_2$ score & APR score\\
 \hline
$\alpha = 0.5$, $\beta = 0.5$ & 53.42 & 49.85 & \textbf{99.93} & 51.77 & 52.57 \\
 \hline
$\alpha = 0.4$, $\beta = 0.6$ & 54.57 & 55.85 & 99.91 & 55.47 & 54.34 \\
 \hline
$\alpha = 0.3$, $\beta = 0.7$ & \textbf{56.42} & 56.85 & 99.93 & \textbf{57.32} & \textbf{56.04} \\
 \hline
$\alpha = 0.2$, $\beta = 0.8$ & 48.57 & 61.00 & 99.89 & 54.53 & 53.31 \\
 \hline
$\alpha = 0.1$, $\beta = 0.9$ & 46.42 & \textbf{65.57} & 99.87 & 56.11  & 51.65 \\
 \hline
\end{tabular}
\caption{Performance metrics (on the test set) for different values of the hyperparameters $\alpha$ and $\beta$ used in training the FCN. The best values for each metric have been highlighted in bold. As expected, it is observed that higher $\beta$ led to higher sensitivity (recall) and lower specificity. The combined performance metrics, in particular APR, $F_2$ and DSC indicate that the best performance was achieved at $\beta=0.7$. Note that the FCN trained with the Dice loss function ($\beta=0.5$) did not generate good results.}
\label{table:res}
\end{table*}
\vspace{-0.3cm}

\begin{figure}
    \centering
    \includegraphics[width=\textwidth]{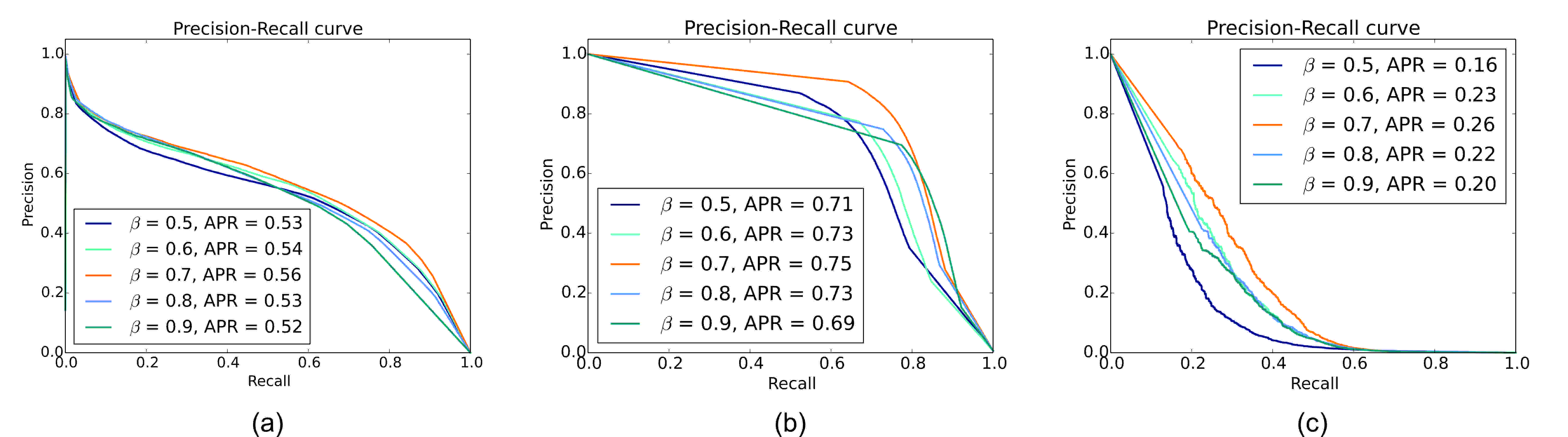}
    \caption{PR curves with different $\alpha$ and $\beta$ for: (a) all test set; (b) a subject with high density of lesions (Fig.~\ref{fig:Tverskey}); and (c) a subject with very low density of lesions (Fig.~\ref{fig:Tverskey2}).}
    \label{fig:PRCurves}
\end{figure}

Figure~\ref{fig:PRCurves}(a) shows the PR curve for the entire test dataset, and Figure~\ref{fig:PRCurves}(b) and (c) show the PR curves for two cases with extremely high, and extremely low density of lesions, respectively. The best results based on the precision-recall trade-off were always obtained at $\beta=0.7$ and not with the Dice loss function. 

\begin{figure}
    \centering
    \includegraphics[width=.9\textwidth]{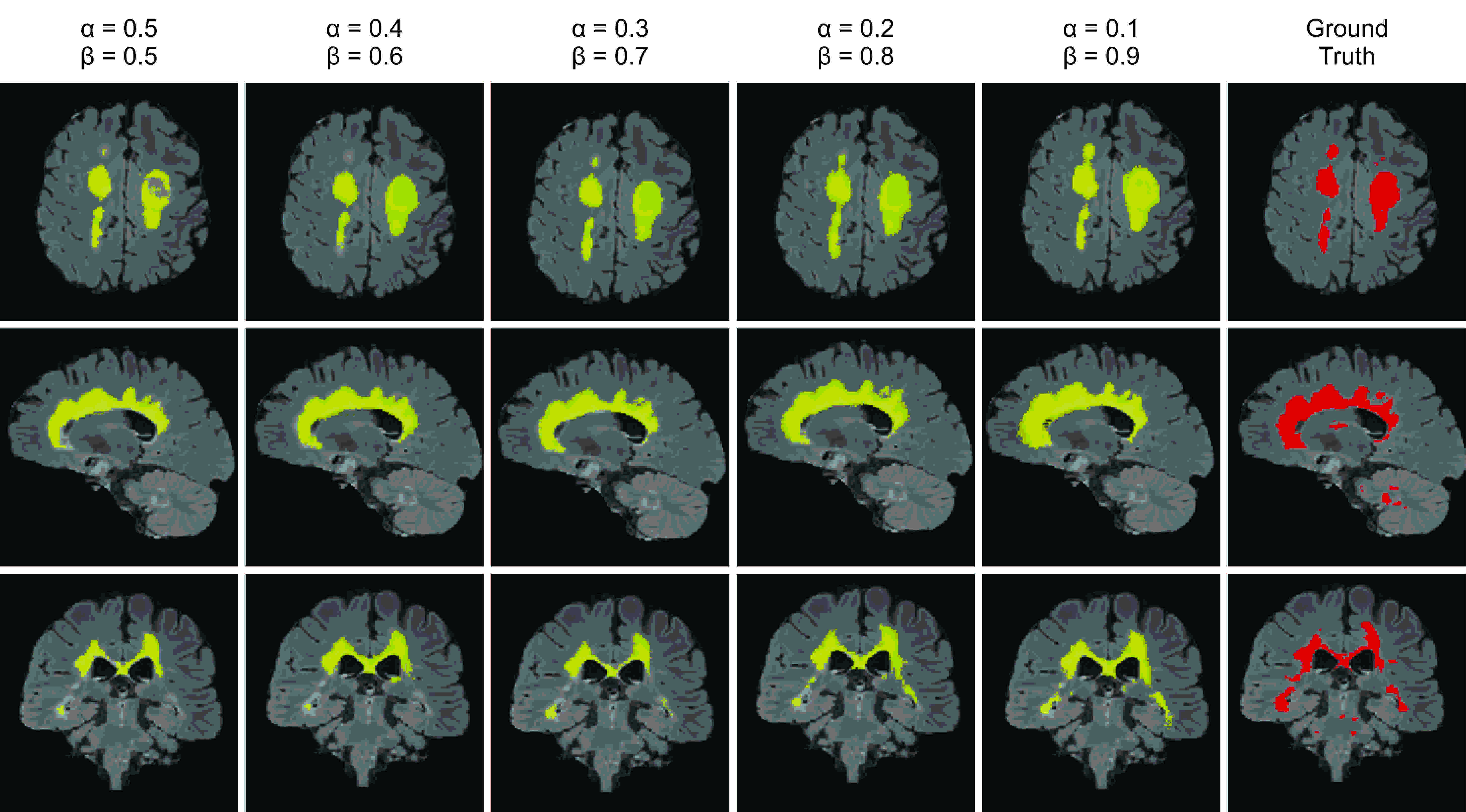}
    \caption{The effect of different penalties on FP and FN in the Tverskey loss function on a case with extremely high density of lesions. The best results were obtained at $\beta=0.7$}
    \label{fig:Tverskey}
\end{figure}

\begin{figure}
    \centering
    \includegraphics[width=.9\textwidth]{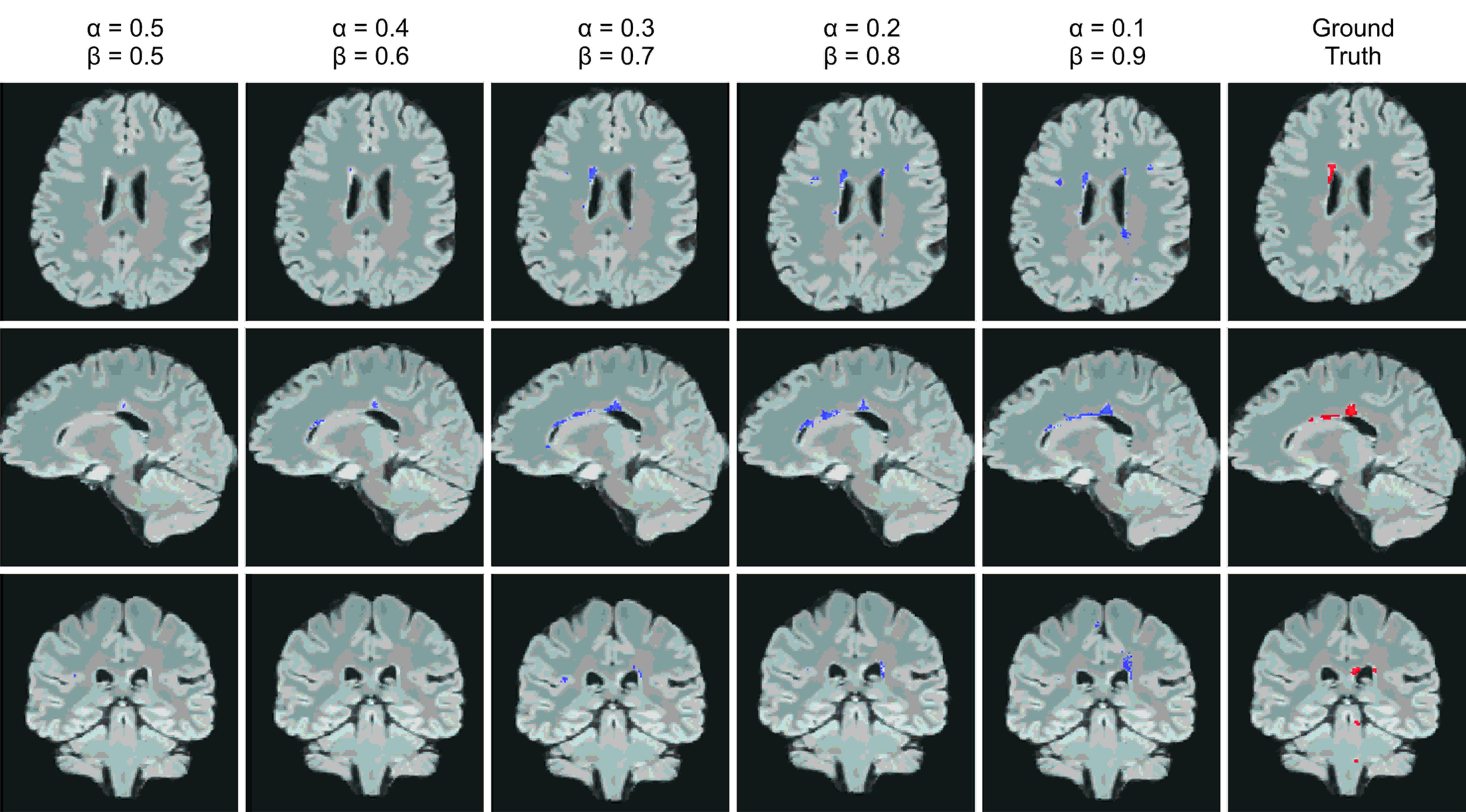}
    \caption{The effect of different penalties on FP and FN in the Tverskey loss function on a case with extremely low density of lesions. The best results were obtained at $\beta=0.7$.}
    \label{fig:Tverskey2}
\end{figure}

Figures~\ref{fig:Tverskey} and~\ref{fig:Tverskey2} show the effect of different penalty magnitudes ($\beta$s) on segmenting a subject with high density of lesions, and a subject with very few lesions, respectively. These cases, that correspond to the PR curves shown in Figure~\ref{fig:PRCurves}(b and c), show that the best performance was achieved by using a loss function with $\beta=0.7$ in training. We note that the network trained with the Dice loss layer ($\beta=0.5$) did not detect the lesions in the case shown in Figure~\ref{fig:Tverskey2}.

\section{Discussion and conclusion}
We introduced a new loss function based on the Tversky index, that generalizes the Dice coefficient and $F_\beta$ scores, to achieve improved trade-off between precision and recall in segmenting highly unbalanced data via deep learning. To this end, we added our proposed loss layer to a state-of-the-art 3D fully convolutional deep neural network based on the U-net architecture~\cite{ronneberger2015u,salehi2017auto}. Experimental results in MS lesion segmentation show that all performance evaluation metrics (on the test data) improved by using the Tversky loss function rather than using the Dice similarity coefficient in the loss layer. While the loss function was deliberately designed to weigh recall higher than precision (at $\beta=0.7$), consistent improvements in all test performance metrics including DSC and $F_2$ scores on the test set indicate improved generalization through this type of training. Compared to DSC which weighs recall and precision equally, and the ROC analysis, we consider the area under the PR curves (APR, shown in Figure~\ref{fig:PRCurves}) the most reliable performance metric for such highly skewed data~\cite{fawcett2006introduction,boyd2013area}. To put the work in context, we reported average DSC, $F_2$, and APR scores (equal to 56.4, 57.3, and 56.0, respectively), which indicate that our approach performed very well compared to the latest results in MS lesion segmentation~\cite{commowick2016msseg,valverde2017improving}. We did not conduct a direct comparison in the application domain, however, as this paper intended to provide proof-of-concept on the effect and usefulness of the Tverky loss layer (and $F_\beta$ scores) in deep learning. Future work involves training and testing on larger, standard datasets in multiple applications to compare against state-of-the-art segmentations using appropriate performance criteria.

\section*{Acknowledgements}
This work was in part supported by the National Institutes of Health (NIH) under grant R01 EB018988. The content of this work is solely the responsibility
of the authors and does not necessarily represent the official views of the NIH.

\bibliographystyle{splncs03}
\bibliography{mybib}

\end{document}